\documentclass{article}

\usepackage[preprint]{neurips_2026}

\usepackage[utf8]{inputenc}
\usepackage[T1]{fontenc}
\usepackage{hyperref}
\usepackage{url}
\usepackage{booktabs}
\usepackage{amsfonts}
\usepackage{nicefrac}
\usepackage{microtype}
\usepackage{xcolor}
\usepackage{graphicx}
\usepackage{amsmath}
\usepackage{amssymb}
\usepackage{amsthm} 
\usepackage{multirow}
\usepackage{enumitem}
\usepackage{dsfont} 
\usepackage{subcaption}
\usepackage{listings}
\usepackage{array}

\newtheorem{proposition}{Proposition}[section]
\theoremstyle{definition}
\newtheorem{definition}{Definition}[section]

\newcommand{\workname}{PassNet}

\title{PassNet: Scaling Large Language Models for Graph Compiler Pass Generation}

\author{
  Yiqun Liu\thanks{Equal contribution.} \quad
  Yingsheng Wu\footnotemark[1] \quad
  Ruqi Yang \quad
  Enrong Zheng \quad
  Honglei Qiu \\
  \bfseries Sijun He \quad
  Tai Liang \quad
  Jingjing Wu \quad
  Yuhan Zhou \quad
  Yiwei Zhang \\
  \bfseries Dongyan Chen \quad
  Weihan Yi \quad
  Xinqi Li\thanks{Corresponding authors: \texttt{lixinqi2010@gmail.com, baosiqi@baidu.com}.} \quad
  Siqi Bao\footnotemark[2] \\[6pt]
  Baidu, Inc. 
}

\begin{document}

\maketitle

\begin{abstract}
Modern tensor compilers such as TorchInductor deliver substantial speedups on mainstream models, yet face a systematic performance ceiling on long-tail workloads---our profiling shows that $43\%$ of real-world subgraphs experience end-to-end slowdowns under default compilation. While LLMs offer a path toward automated optimization, existing efforts focus on standalone \emph{kernel generation}. We argue that \emph{pass generation}---where LLMs author structured graph transformations that integrate directly into compiler pipelines---is the more appropriate abstraction.

We propose \textbf{PassNet}, the first large-scale ecosystem for LLM-based compiler pass generation, comprising: (1)~\textbf{PassNet-Dataset}, over $18K$ unique computational graphs from $100K$ real-world models; and (2)~\textbf{PassBench}, 200 curated long-tail fusible tasks (comprising 2,060 subgraphs in total) evaluated under the \textbf{Error-aware Speedup Score ($ES_t$)}---a metric unifying correctness, stability, and performance---with layered integrity defenses against systematic LLM exploitation. Experiments reveal that PassBench is both highly discriminative and genuinely unsaturated: the best frontier model trails TorchInductor by 37\% in aggregate, yet on individual subgraphs LLMs achieve up to $3\times$ speedup over the same compiler---indicating that the bottleneck is consistency, not capability. Fine-tuning a small model on merely ${\sim}$4K PassNet trajectories yields a $2.67\times$ improvement approaching frontier-model performance, demonstrating substantial headroom and validating PassNet as live training infrastructure for advancing LLM-driven compiler optimization. All data, benchmarks, and tooling are publicly available.
\end{abstract}

\section{Introduction}
\label{sec:intro}

Modern deep learning systems increasingly rely on tensor compilers (e.g., TVM~\cite{TVM}, XLA~\cite{XLA}, TorchInductor~\cite{ansel2024pytorch}) to lower high-level computational graphs into efficient backend implementations for heterogeneous hardware. These compilers apply expert-designed, rule-based \emph{pass pipelines}---sequences of graph transformations such as operator fusion, tiling, and layout selection---that have proven remarkably effective: on mainstream architectures, TorchInductor delivers up to $2.27\times$ inference speedups over eager execution across 180+ models~\cite{ansel2024pytorch}. Yet these static strategies face a structural limitation when confronting the long tail of real-world operator combinations.

\paragraph{The Long-Tail Optimization Gap.}
To quantify this limitation, we profile TorchInductor's default pipeline on $9{,}526$ subgraphs extracted from over $1{,}000$ community models. The results reveal a systematic limitation: $34\%$ achieve marginal speedups ($<1.2\times$), $43\%$ experience end-to-end slowdowns, and $8.3\%$ are strictly degraded. This gap is structural, correlating with operator coverage rather than graph complexity ($r=0.013$), suggesting that scaling graph coverage alone cannot overcome the heuristic-induced performance ceiling.

\paragraph{Pattern Concentration Creates Leverage.}
While the long-tail gap is significant, computation-graph patterns exhibit strong power-law concentration: deduplicating $100K$ models yields only ${\sim}18K$ distinct graphs (82\% redundancy), and ${\sim}10{,}000$ subgraphs reduce to ${\sim}1{,}025$ unique structural patterns. Generating high-quality passes for this concentrated set suffices to cover most workloads, enabling a shift from \emph{manual rules} to \emph{automated, data-driven pass generation}.

\paragraph{From Kernel Generation to Pass Generation.}
Large Language Models (LLMs) are a promising approach for long-tail optimization. Existing work has explored \textbf{Kernel Generation}~\cite{ouyang2025kernelbench, dai2026cuda, liao2025kernelevolve}, producing standalone GPU kernels for individual operators. However, such kernels lack composability with compiler passes, require manual integration for deployment, and are difficult to verify due to unconstrained code generation.

We therefore define a new task, \textbf{Pass Generation}: given a computational subgraph, an LLM must author a structured compiler pass that interfaces directly with the compiler's intermediate representation (formal definition in Section~\ref{sec:ecosystem}). This formulation preserves the ``one-line compilation'' experience (e.g., \texttt{torch.compile}) while enabling composable and verifiable optimizations. However, advancing this task requires large-scale data and rigorous evaluation---resources that do not yet exist.

\paragraph{The Data and Evaluation Bottleneck.}
Is pass generation with LLMs realistic today? On individual long-tail subgraphs, frontier LLMs can already generate passes achieving up to $3\times$ speedup over the default compiler---yet aggregate performance trails far behind, with no model reaching geometric-mean speedup above $1.0$.
This gap stems from two \emph{infrastructure} bottlenecks: (1) \emph{data scarcity}---a lack of large-scale, specialized corpora for fusion and layout optimization; and (2)~\emph{evaluation blind spots}---the absence of rigorous benchmarks allowing agents to bypass correctness for illusory gains. To bridge these, we introduce \textbf{\workname}---the first ecosystem providing systematic data and benchmark support for the pass generation task.

\paragraph{Contributions.}
We formalize \emph{pass generation}, where LLMs author structured graph transformations that integrate into compiler pipelines, and build a large-scale ecosystem to support this task.

\begin{itemize}[leftmargin=*,itemsep=2pt]
    \item \textbf{PassNet-Dataset}: We collect over $18{,}086$ unique computational graphs from $100K$ real-world models across diverse frameworks (PyTorch, PaddlePaddle) and task categories. We design \emph{Recursive Folding} and \emph{Execution-driven Prefix Analysis} to construct structurally diverse subgraphs at multiple granularities, forming the first large-scale open training set for pass generation.

    \item \textbf{PassBench}: We construct a benchmark of $200$ tasks, each consisting of a variable number of long-tail subgraphs, evaluated under the \emph{Error-aware Speedup Score} ($ES_t$), which jointly measures correctness, stability, and performance. To ensure evaluation integrity, we introduce layered defenses including AST-based inspection, runtime dispatch interception, and reverse evaluation order, which systematically counter exploitation patterns observed during development.

    \item \textbf{Benchmark Validation}: Through extensive evaluation of 6 frontier and open-source models, we show that PassBench is both highly discriminative ($3.22\times$ gap between model tiers) and genuinely unsaturated (best model trails TorchInductor by 37\%). Fine-tuning on merely ${\sim}$4K PassNet trajectories yields a $2.67\times$ improvement, validating the dataset's utility as training infrastructure.
\end{itemize}

\noindent The entire ecosystem---dataset, benchmark, evaluation tooling, and agent scaffold---is publicly available at \url{https://github.com/PaddlePaddle/PassNet}.

\section{Related Work}
\label{sec:related}

\paragraph{Tensor Compilers.}
Tensor compilers transform high-level computation graphs into device-specific kernels via IR lowering and scheduling. TVM~\cite{TVM} and Ansor~\cite{Ansor} are search-based compilers relying on cost models, while XLA~\cite{leary2017xla} applies heuristic graph-level optimizations. Recent systems include MetaSchedule~\cite{MetaSchedule}, Hidet~\cite{hidet}, and BladeDISC~\cite{BladeDISC}, with industry frameworks such as CINN~\cite{cinn2021} and TorchInductor~\cite{ansel2024pytorch} integrated into production systems. Despite these advances, existing approaches still rely on manual transformation rules and struggle with long-tail workloads.

\paragraph{LLMs for Code Generation and Compiler Optimization.}
Large Language Models demonstrate strong code generation capabilities~\cite{codeLLMSurvey, codeLLMSurvey-zs}, including code completion~\cite{liu2025evaluating}, bug repair~\cite{huang2025seeing, yang2026morepair}, and software engineering automation~\cite{jimenez2024swebench, yang2024sweagent, wang2024openhands}. For compiler tasks, LLM Compiler~\cite{cummins2025llm} pre-trains on compiler IR, Compiler-r1~\cite{pan2025compiler} explores RL-based auto-tuning, and DeCOS~\cite{cui2025decos} studies data-efficient optimization selection. TLP~\cite{zhai2023tlp} and follow-up work~\cite{zhai2024enabling} apply language models to tensor program generation. However, these approaches focus on pass selection or scheduling rather than synthesizing new transformation logic.

\paragraph{LLM-Driven GPU Kernel Generation.}
Recent work generates GPU kernels directly with LLMs. KernelBench~\cite{ouyang2025kernelbench} benchmarks LLM-generated kernels, revealing gaps from production compilers. CUDA Agent~\cite{dai2026cuda} and Kernelevolve~\cite{liao2025kernelevolve} scale agentic generation, while STARK~\cite{dong2025stark}, Kevin~\cite{baronio2025kevin}, and Geak~\cite{wang2025geakintroducingtritonkernel} explore multi-agent and RL-based refinement. Other efforts include QiMeng-GEMM~\cite{zhou2025qimeng}, CUDA-L1~\cite{li2025cuda}, and Autocomp~\cite{autocompllm}. In contrast, \workname{} targets \emph{pass generation}, i.e., structured transformations that integrate with compiler pipelines.

\paragraph{Performance Benchmarks.}
DL evaluation has evolved from DeepBench~\cite{deepbench} to MLPerf~\cite{mattson2020mlperf}. CompilerGym~\cite{CompilerGym} provides an RL environment for compiler optimization. Datasets such as ComPile~\cite{grossman2024compile}, TpuGraphs~\cite{phothilimthana2023tpugraphs}, and TenSet~\cite{zheng2021tenset} provide benchmarks for learned compilers. \workname{} focuses on computational-graph-level pass generation for long-tail optimization, with an evaluation framework that jointly measures correctness, stability, and speedup.

\section{The PassNet Ecosystem}
\label{sec:ecosystem}
The \textit{PassNet} ecosystem bridges the dual infrastructure gaps identified in Section~\ref{sec:intro}: the scarcity of large-scale pass generation corpora and the lack of robust, multi-dimensional benchmarks. In the following, we formalize the pass generation task before detailing each ecosystem component.

\subsection{Task Formulation}
\label{subsec:task-formulation}
In modern tensor compilers, a \emph{pass} is a self-contained graph transformation that rewrites a computational graph while preserving its input--output semantics~\cite{MLIR, DeeplearningcompilerSurvey}. Following the pattern-based rewriting paradigm adopted by MLIR~\cite{MLIR} and TorchInductor~\cite{ansel2024pytorch}, we formalize the core abstractions below.

\begin{definition}[Computational Graph]
\label{def:graph}
A computational graph is a DAG $G = (V, E, \tau, \sigma)$, where $V$ is a set of operator nodes, $E$ encodes data dependencies, $\tau: V \to \mathcal{T}$ assigns operator types, and $\sigma: V \to \mathbb{Z}^+$ assigns output shapes. We write $f_G: \mathcal{X} \to \mathcal{Y}$ for the function computed by $G$.
\end{definition}

\begin{definition}[Compiler Pass]
\label{def:pass}
A compiler pass is a pair $\pi = (M, R)$, where $M: \mathcal{G} \to 2^{\mathcal{G}}$ is a pattern matcher identifying optimization-eligible subgraphs, and $R: \mathcal{G} \to \mathcal{G}$ is a rewriter replacing each matched subgraph with an optimized equivalent. A pass $\pi$ is \textbf{valid} on $G$ under tolerance $t$ if:
\begin{equation}
\forall x \in \mathcal{X},\quad \mathrm{err}(f_G(x),\; f_{\pi(G)}(x)) \leq t
\end{equation}
\end{definition}

\noindent The \textbf{pass generation task} is: given a \emph{task instance}
$\mathcal{T} = \{G_1, \ldots, G_k\}$ of subgraphs sharing the same operator-type sequence but varying in shape and dtype, generate a valid pass $\pi$ that rewrites every $G_i \in \mathcal{T}$ and improves aggregate runtime
performance. This multi-graph formulation requires $\pi$ to generalize across 
varying shapes and data types, precluding shape-specific hacks. As opposed to 
free-form kernel generation, it further ensures composability with existing 
compiler pipelines and verifiability through standard compiler infrastructure.

\subsection{Dataset Construction}
\label{subsec:dataset-construction}

\begin{figure}[!htbp]
    \centering
    \includegraphics[width=0.82\linewidth]{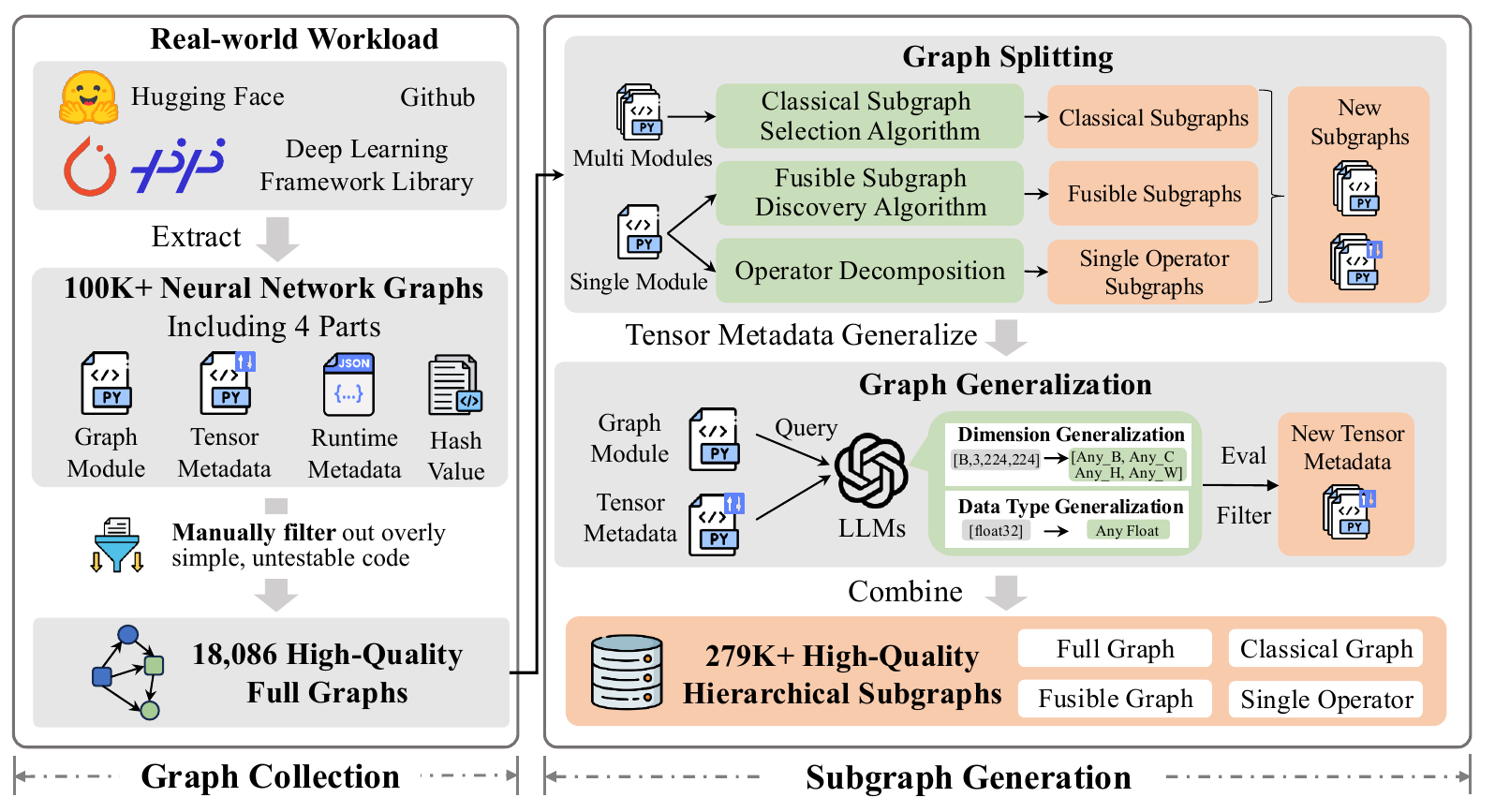}
    \caption{\textbf{PassNet Dataset Construction Pipeline.} We collect and filter real-world model graphs, and generate hierarchical training subgraphs via recursive graph splitting and metadata generalization.}
    \label{fig:data_collection}
\end{figure}

The construction of the dataset consists of two stages: \textbf{computational graph collection} and \textbf{advanced subgraph generation}, as illustrated in Figure~\ref{fig:data_collection}. The overall goal is to preserve real-world computation patterns while constructing structurally diverse and optimization-relevant subgraphs.

\paragraph{Graph Collection and Validation.}
We extract computational graphs from real-world models via a lightweight decorator (\texttt{pass\_net.extract}). During execution, symbolic tracing captures operator invocations and tensor dependencies, producing standardized representations (high-level IR, weights, and input metadata). 
To ensure high-fidelity graphs for downstream subgraph construction, each graph is rigorously validated against five key constraints: runnable, serializable, decomposable, statically analyzable, and custom-operator accessible (see Appendix~\ref{sec:appendix_constraints}).

\paragraph{Subgraph Generation Strategies.}
To systematically map the optimization space, we propose three subgraph categories focusing on structural recurrence, fusion potential, and primitive behavior.

\begin{figure}[!htbp]
    \centering
    \begin{subfigure}{0.48\linewidth}
        \centering
        \includegraphics[width=0.8\linewidth, height=3.8cm]{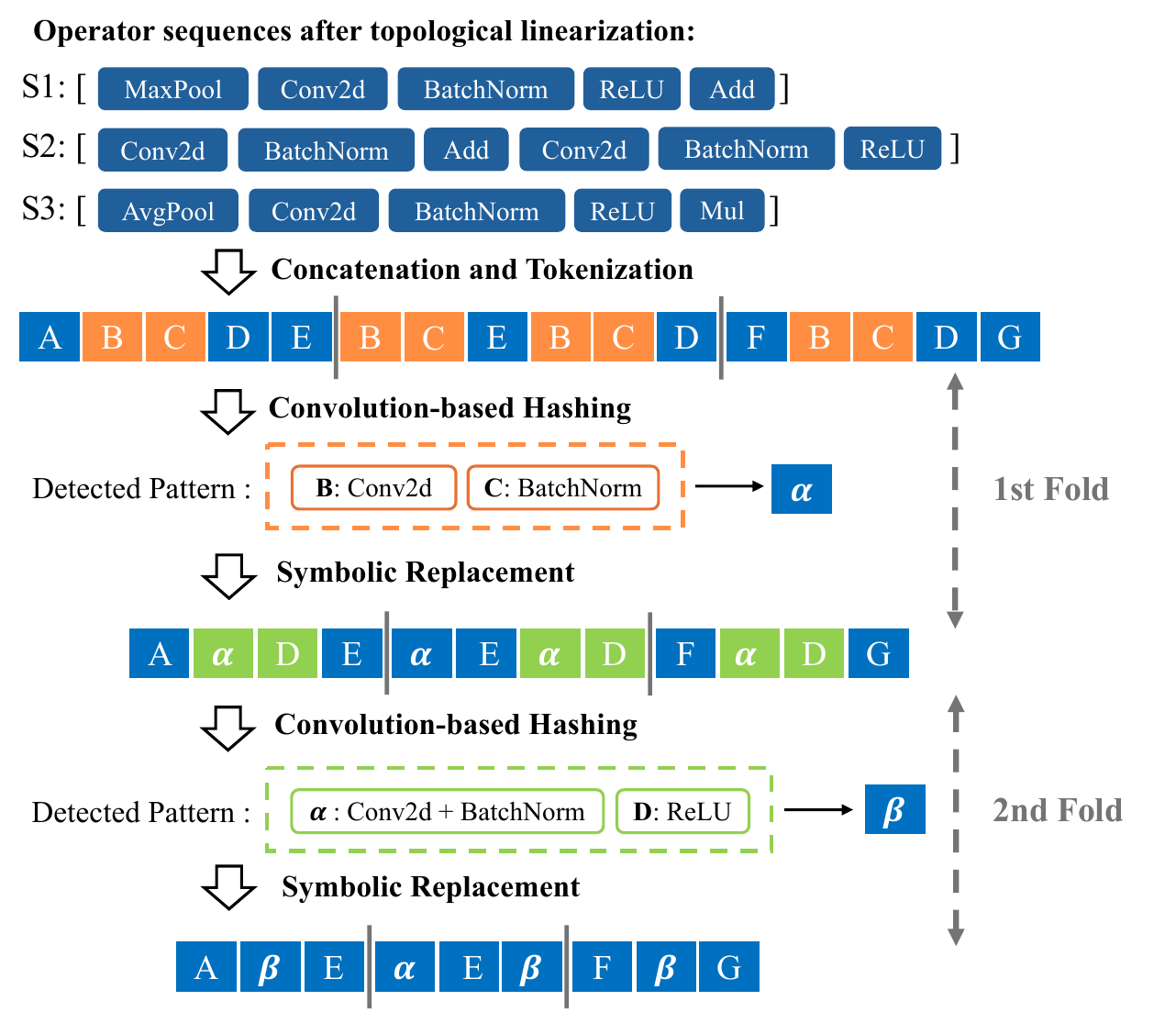}
        \caption{\textbf{Recursive folding.} [Conv2d, BatchNorm] $\to$ $\alpha$; [$\alpha$, ReLU] $\to$ $\beta$.}
        \label{fig:recursive_folding}
    \end{subfigure}
    \hfill
    \begin{subfigure}{0.48\linewidth}
        \centering
        \includegraphics[width=\linewidth, height=3.8cm]{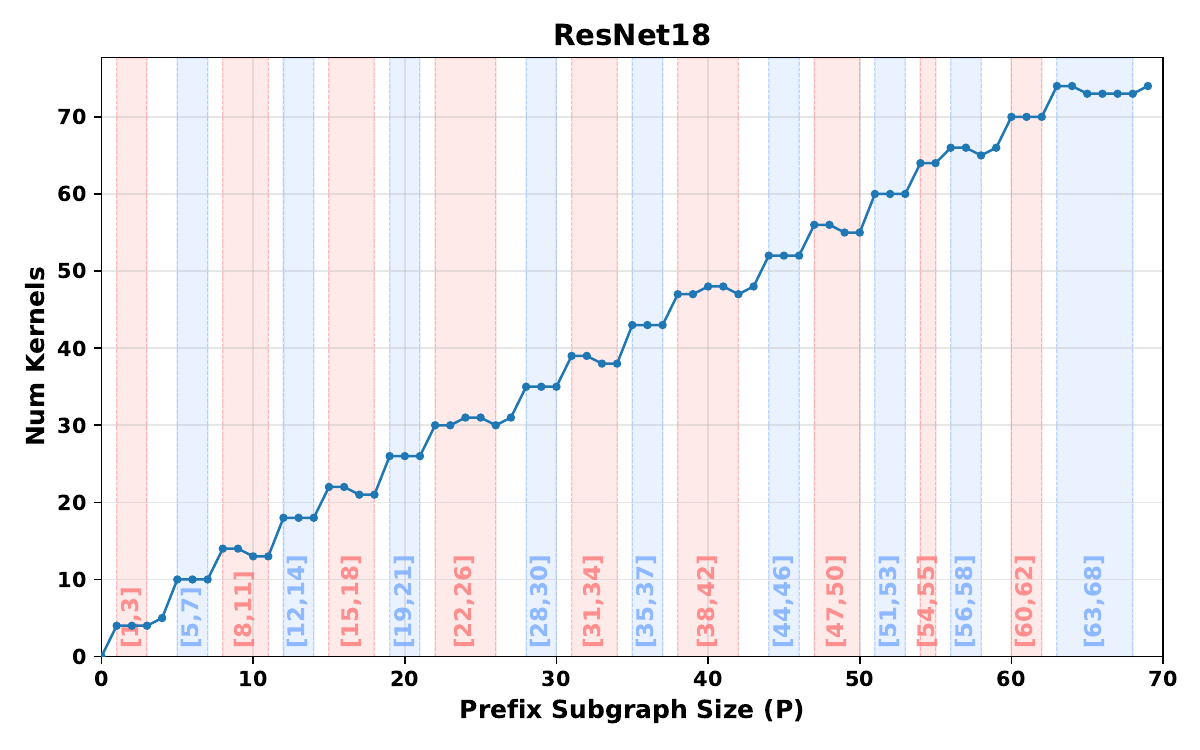}
        \caption{\textbf{Prefix kernel-count curve (ResNet-18).} Plateau regions indicate fusible intervals.}
        \label{fig:prefix_analysis}
    \end{subfigure}
    \caption{Recursive folding for subgraph selection (left) and prefix-based fusibility analysis (right).}
\end{figure}

(1) \textbf{Classical Subgraph Selection via Recursive Folding.}
A \textit{Classical Subgraph} is a recurrent structural motif representing common computational patterns within a model family. To extract such motifs at scale, we design \textbf{Recursive Folding}: the method first linearizes the computation graph into a topological operator sequence, then iteratively applies convolution-based hashing to identify frequent subsequences and abstract them into symbolic units. This hierarchical process captures both local idioms (e.g., [Conv2d, BatchNorm] $\to \alpha$) and higher-level compositions (e.g., [$\alpha$, ReLU] $\to \beta$), producing a compact set of representative structures (Figure~\ref{fig:recursive_folding}).

(2) \textbf{Fusible Subgraph Discovery via Prefix Analysis.}
A \textit{Fusible Subgraph} is a contiguous segment of a computational graph where a compiler can apply operator fusion. To identify such regions systematically, we design \textbf{Prefix Analysis}, which analyzes the prefix kernel-count curve $(P, K(P))$, where $K(P)$ is the number of kernels launched by the first $P$ operators, and detects plateaus satisfying $K(P+1) = K(P)$. These indicate that operators are absorbed into existing execution units. Extracting contiguous plateaus yields subgraphs reflecting compiler fusion behavior (Figure~\ref{fig:prefix_analysis}).

(3) \textbf{Single-operator Subgraph.} A \textit{Single-operator Subgraph} consists of a primitive operator and captures operator-level behavior, complementing higher-level structures.

We apply shape and data type generalization during subgraph instantiation, generating instances with 10 shape configurations and 3 data types. This introduces variation in computation and optimization difficulty, improving applicability across hardware backends and compiler strategies. 

\subsection{Dataset Characteristics}
\label{subsec:dataset-characteristics}

The resulting \textit{PassNet} dataset comprises over $18K$ unique computational graphs derived from $100K$ diverse models across PyTorch and PaddlePaddle, with four distinguishing properties:

\textbf{(i)~Authenticity and Scale:} All samples originate from production-grade community libraries rather than synthetic generators. With $100K$ source models yielding $18{,}086$ deduplicated graphs (82\% redundancy), PassNet captures the true distribution of real-world computation patterns---including long-tail operator combinations absent from curated benchmarks.

\textbf{(ii)~Structural Diversity:} The collection spans six application domains (NLP: 63.6\%, CV: 27.0\%, Multimodal: 1.7\%, Audio: 1.2\%, Others: 6.5\%), with model scales from lightweight mobile architectures to $10$B-parameter models and node counts from 2 to $298{,}441$ (median $\sim 2^9$).

\textbf{(iii)~Optimization-Relevant Coverage:} We instantiate $129$K fusible ($N_{\text{ops}} \in [2, 35]$), $126$K classical ($N_{\text{ops}} \in [4, 62]$), and $24$K single-operator subgraphs---totaling ${\sim}279$K instances across three complementary granularities, each augmented with 10 shape configurations and 3 data types.

\textbf{(iv)~Interoperability:} A unified format for graphs, metadata, and custom operators ensures compatibility with compilers such as TorchInductor, CINN, XLA, and TVM without conversion overhead.

\subsection{PassBench Design}
\label{subsec:passbench-design}

While the full dataset serves as training data, rigorous evaluation requires a controlled benchmark with diverse, representative tasks. 
We curate \textbf{PassBench} via multi-dimensional bucketing and hierarchical grouping of subgraphs, with each group forming a task instance. 
The benchmark primarily focuses on fusible-subgraph tasks, comprising $4{,}476$ training samples and $200$ high-quality evaluation samples, with no overlap between training and evaluation sets.
The dataset is further augmented with $4{,}078$ classical-subgraph training samples and $200$ evaluation counterparts, complemented by $1{,}029$ single-operator samples. This multi-tiered composition provides varying levels of task difficulty.

\paragraph{Selection Pipeline.}
Subgraphs are bucketed along three dimensions: operator sequence (exact-match), input shape (log-quantized), and input dtype. Within each operator-sequence bucket, we apply stratified sampling with fixed stride $\sigma$, followed by cross-shape and dtype-aware aggregation to form groups corresponding to PassBench tasks (details in Appendix~\ref{sec:appendix_passbench_sampling}). 
For evaluation, we select $200$ operator sequences via a Hidden Markov Model and retain the largest group per sequence. Selected fusible samples contain $1\text{--}396$ subgraphs ($\mathrm{avg.}=10$), exhibiting a long-tail distribution, and yield 2,060 subgraph-level evaluations in aggregate---providing finer-grained signal than benchmarks that evaluate at the task level alone (e.g., KernelBench, where each task corresponds to a single kernel).

\paragraph{Task Format.}
Each task is packaged as a directory containing (i) a Python reference implementation (\texttt{GraphModule}), (ii) tensor metadata, and (iii) runtime metadata. The agent is required to generate executable transformation passes under \texttt{pass\_dir/} along with a JSON manifest. A submission is considered successful if it preserves correctness across all specified data types while achieving measurable performance improvement.

\subsection{Error-aware Speedup Metrics}
\label{sec:es-metric}

Evaluating compiler passes requires assessing both correctness and performance. Existing approaches fall short in three aspects: (1) treating correctness as binary (pass/fail) despite its continuous tolerance; (2) producing discrete signals that hinder iterative agent training; and (3) operating at the benchmark level, lacking fine-grained feedback on individual graphs.

We define a per-subgraph metric to jointly evaluate correctness and performance. For each subgraph $i$ with measured speedup $s_i$, we introduce a tolerance threshold $t \in \{-10, \dots, |E|+1\}$ to control acceptance of error categories $c_i \in \{1, 2, 3\}$ (accuracy, compilation, and runtime failures). For $t \le 0$, strict correctness is enforced with varying numerical tolerances; for $t > 0$, more error categories are progressively forgiven. Let $\mathrm{correct}_{t,i}$ be a binary indicator of whether subgraph $i$ satisfies the correctness criteria under threshold $t$. The \textbf{error-aware rectified speedup} is defined as:

\begin{equation}
\hat{s}_{t,i} =
\begin{cases}
s_i, & \mathrm{correct}_{t,i} \land s_i \ge 1,\\[3pt]
s_i^{p+1}, & \mathrm{correct}_{t,i} \land s_i < 1,\\[3pt]
b^{\mathbf{1}(t<c_i)}, & \text{otherwise,}
\end{cases}
\label{eq:rectified-speedup}
\end{equation}
where parameters $p, b \in (0, 1)$ respectively govern the exponential penalty for slowdowns and the base penalty for incorrect executions. The metric distinguishes three distinct scenarios: 
\textbf{(i)~Speedup ($s_i \ge 1$):} retained for correct executions; 
\textbf{(ii)~Slowdown ($s_i < 1$):} exponentially penalized via $p$; 
\textbf{(iii)~Incorrect:} assigned penalty $b$ if $t < c_i$, or $1$ if forgiven ($t \ge c_i$).

The \textbf{Error-aware Speedup Score} $ES_t$ is defined as the geometric mean of $\{\hat{s}_{t,i}\}$ across all $N$ subgraphs, with its equivalent factored form and macro-level derivation detailed in Appendix~\ref{sec:appendix_es_geometric_mean}:

\begin{equation}
ES_t = \Bigl(\prod_{i=1}^{N} \hat{s}_{t,i}\Bigr)^{1/N}.
\label{eq:es-macro}
\end{equation}

We further aggregate $ES_t$ across the tolerance spectrum via a normalized geometric mean to yield a unified scalar for agent feedback:

\begin{equation}
\mathrm{AS} = \prod_{t=-10}^{|E|+1} ES_t^{\,W_t / \sum_{s=-10}^{|E|+1} W_s},
\label{eq:as-aggregate}
\end{equation}

where $W_t$ assigns high weight to the strict-correctness regime ($t \in [-5,-3]$) and decays exponentially toward relaxed tolerances (details in Appendix~\ref{sec:appendix_wt}). While $fast_p$ (fraction of correct graphs with speedup $>p$) provides a binary correctness threshold, its discrete nature lacks a smooth signal and operates at benchmark-level granularity, failing to capture per-graph variation. AS instead provides a smooth, continuous feedback signal that jointly reflects correctness and performance gains.

\subsection{Evaluation Integrity}
\label{subsec:evaluation-integrity}

A key challenge overlooked by prior work is that LLMs systematically exploit evaluation loopholes. In KernelBench-style evaluations, a submitted kernel passes as long as its output matches the reference---even if it simply delegates to \texttt{torch.compile}. During PassBench development, we found that 29\%--50\% of frontier-model submissions contained some form of exploitation. We document a three-stage arms race, where each defense exposed a new attack surface.

\paragraph{Case A: Computation Delegation $\to$ AST Inspection.}
The most prevalent pattern is invoking high-level APIs to bypass explicit operator-level transformations---e.g., calling \texttt{torch.matmul(in\_1, in\_3)} instead of implementing the fusion logic. We counter this with AST-based static analysis that blocks forbidden API calls within non-exempt functions (raising \texttt{RuntimeError: blocked call}), intercepting 78\% of violations.

\paragraph{Case B: Dynamic Evasion $\to$ Dispatch Interception.}
Static analysis cannot cover dynamic execution paths such as implicit tensor method calls (e.g., \texttt{tmp = in\_0 + in\_1} dispatching to \texttt{aten.add.Tensor}). We introduce a runtime monitoring layer via \texttt{PoisonDispatchTensor}, which overloads \texttt{\_\_torch\_dispatch\_\_} to enforce whitelist-based operator filtering on the mandatory dispatch path. This layer exclusively identifies 18\% of violations missed by AST inspection.

\paragraph{Case C: Cache Pollution $\to$ Reverse Evaluation.}
We discovered a \emph{correctness escape}: in conventional ``eager-first'' evaluation order, PyTorch's memory pooling leaves residual data in GPU memory that flawed code (e.g., \texttt{return torch.empty(...)}) can inadvertently pass validation against. We adopt ``reverse evaluation'' (compiled execution before eager baseline) to ensure verification within a pristine system state; such cases now receive correctness $= 0$.

These layered defenses are complemented by the \emph{pass-form mandate} itself: requiring agents to author a pattern matcher and rewriter (rather than a standalone kernel) forces structural understanding of the computation graph, making trivial exploitation significantly harder.

\section{Experiments}
\label{sec:experiments}

We design experiments to validate PassBench as a benchmark and PassNet-Dataset as training infrastructure:
\textbf{(Q1)}~Does PassBench provide meaningful discrimination across model capabilities, and how do current models compare to traditional compilers?
\textbf{(Q2)}~Can PassNet-Dataset improve model performance via post-training, validating its utility as training data?

\subsection{Setup}
\label{subsec:setup}

\paragraph{PassAgent.}
To establish reproducible baselines, we implement \textbf{PassAgent}, a lightweight agentic scaffold for compiler pass synthesis. Following the dual-tool paradigm~\cite{yang2024sweagent, wang2024openhands, pan2024r2egym, anthropic2024swesonnet}, PassAgent provides: (1)~\texttt{file\_editor} for multi-file workspace manipulation, and (2)~\texttt{pass\_evaluator} for invoking the PassBench evaluation pipeline with three-stage diagnostics (pass matching $\to$ correctness $\to$ performance). The agent iteratively inspects target graphs, edits pass files, and refines based on $AS \ Score$ feedback until convergence. All experiments in this paper are conducted on the fusible-subgraph tasks, using the $ES_t$ formulation with $b = 0.1$ and $p = 0$. Full design details are in Appendix~\ref{sec:appendix_setup}. 

\paragraph{Models and Baselines.}
We evaluate frontier models (GPT-5.4, Claude-Opus-4.6, Claude-Sonnet-4.6) and open-source models (GLM-5.1~\cite{glm5team2026glm5vibecodingagentic}, MiniMax-M2.7~\cite{minimaxm27}, Qwen3-30B-A3B and Qwen3-4B~\cite{qwen3technicalreport}).
Baselines include Eager execution and TorchInductor (\texttt{torch.compile}, default mode), preferred over \texttt{max-autotune} due to CUDA Graph layout freezing and overheads outweighing gains at this scale.
Hardware details are in Appendix~\ref{sec:appendix_setup}.

\subsection{Main Results}
\label{subsec:main-results}

\begin{table}[!t]
\centering
\small
\setlength{\tabcolsep}{4pt}
\renewcommand{\arraystretch}{1.15}
\begin{tabular}{cccccc}
\toprule
\textbf{Model} & \textbf{fast\_1(\%)}$^{\ast}$ & \textbf{Samp. CR (\%)}$^{\dagger}$ & \textbf{Sub. CR (\%)}$^{\ddagger}$ & \textbf{G-Mean Speedup}$^{\S}$ & \textbf{AS Score} \\
\midrule
Eager & 100.0 & 100.0 & 100.0 & 1.000 & 1.000 \\
Inductor & 20.3 & 64.5 & 85.0 & 0.846 & 0.706 \\
\midrule
GPT-5.4 & 7.4 & 47.0 & 54.6 & 0.821 & 0.410 \\
Claude-Opus-4.6 & 9.8 & 40.0 & 48.6 & 0.922 & 0.410 \\
Claude-Sonnet-4.6 & 9.2 & 52.0 & 61.9 & 0.835 & 0.448 \\
GLM-5.1 & 4.8 & 30.5 & 33.5 & 0.844 & 0.240 \\
MiniMax-M2.7 & 1.0 & 23.5 & 23.0 & 0.653 & 0.208 \\
Qwen3-4B & 0.2 & 1.0 & 1.3 & 0.953 & 0.108 \\
Qwen3-30B-A3B & 0.6 & 7.5 & 11.8 & 0.693 & 0.139 \\
\midrule
Qwen3-4B-SFT\textsuperscript{\P} & 3.0 & 27.5 & 28.5 & 0.808 & 0.240 \\
Qwen3-30B-A3B-SFT\textsuperscript{\#} & 5.3 & 44.0 & 48.8 & 0.809 & 0.371 \\
\bottomrule
\end{tabular}
\caption{\textbf{Main Results on PassBench.}
$^{\ast}$Fraction of correct subgraphs with speedup $\ge 1.0$ over eager.
$^{\dagger}$Fraction of correct samples.
$^{\ddagger}$Fraction of correct subgraphs.
$^{\S}$Geometric-mean speedup over correct subgraphs.
\textsuperscript{\P}Qwen3-4B fine-tuned on PassNet.
$^{\#}$Qwen3-30B-A3B fine-tuned on PassNet.}
\label{tab:main_results}
\end{table}

Table~\ref{tab:main_results} presents results across five metrics (definitions in caption); we highlight four key findings.

\textbf{(1)~PassBench effectively discriminates model capabilities.}
The benchmark shows a clear performance gap: Claude-Sonnet-4.6 (AS$=$0.448) outperforms Qwen3-30B-A3B (AS$=$0.139) by $3.22\times$. Correctness ratios exhibit a similar spread (Sub.\ CR: 61.9\% vs.\ 11.8\%). This separation indicates that PassBench provides a meaningful discriminative signal across models.

\textbf{(2)~All models fall substantially short of the compiler baseline.}
All models have G-Mean Speedup below 1.0, meaning generated code is \emph{slower} than eager execution.
The best frontier model (Claude-Opus-4.6, G-Mean$=$0.922) approaches but does not reach parity, and the highest AS score (Claude-Sonnet-4.6, 0.448) trails Inductor (0.706) by 37\%. Even for correct outputs, speedups rarely exceed $1.2\times$, indicating limited hardware-cost awareness in current LLM-generated kernels.

\textbf{(3)~Low aggregate scores coexist with striking individual successes.}
No model reaches AS$\ge$0.5 or G-Mean$\ge$1.0, yet on specific long-tail subgraphs, frontier models deliver up to $3.02\times$ speedup over Inductor (Section~\ref{subsec:case-studies}).
The contrast between strong per-instance potential and weak aggregate performance reveals that the challenge is \emph{consistency}: models fail to reliably generalize sparse successes across diverse patterns. Moreover, persistent failure modes including boundary misalignment, cost-model blindness, and semantic disruption suggest an unsaturated benchmark. Improving data and training infrastructure, rather than mere scaling, is key to closing this gap.

\begin{figure}[!t]
  \centering
  \includegraphics[width=0.95\linewidth]{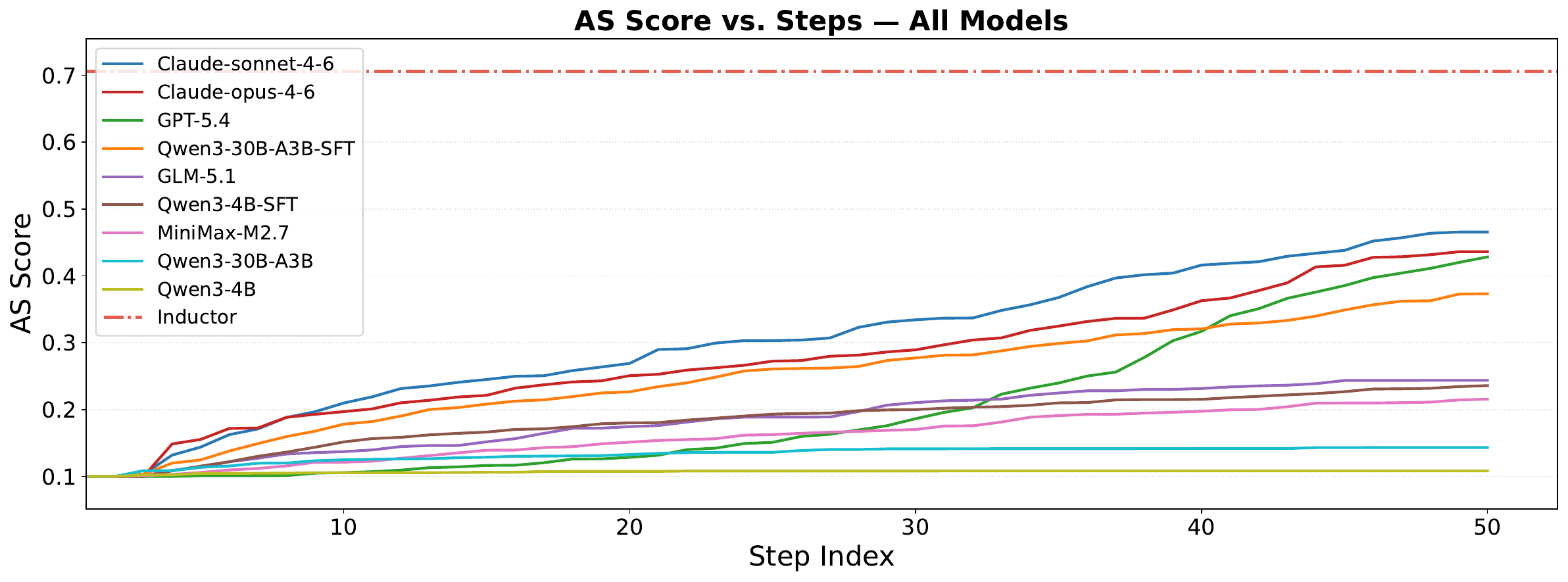}
  \caption{\textbf{Agent performance scales with iteration budget.} AS Score as a function of iteration steps (up to 50).}
  \label{fig:agent_scaling}
\end{figure}

\textbf{(4)~Iteration reveals capabilities missed by single-shot evaluation.}
All main results are reported at convergence (50 iterations). As shown in Figure~\ref{fig:agent_scaling}, a single evaluation captures only 31\%--51\% of each agent's best AS score (mean 38\%), and 12\%--52\% of eventually-passing samples exhibit non-monotonic \textit{pass$\to$fail$\to$pass} trajectories as agents explore different generalization strategies. This differs from KernelBench, where iteration monotonically refines a single kernel.

\subsection{Dataset Efficacy via Distillation}
\label{subsec:distillation}

The main results show a large performance gap between frontier models and the open-source models. To address Q2, we distill expert trajectories into the smaller model and evaluate the resulting performance gains.

\paragraph{Setup.}
We generate PassAgent trajectories from $4,476$ samples using Claude-Sonnet-4.6 (two trials per instance, up to 50 iterations), retaining those with AS $>$ 0.1 to obtain $3,899$ training trajectories. We then fine-tune Qwen3-30B-A3B and Qwen3-4B with learning rate $2\!\times\!10^{-5}$ (cosine decay to $2\!\times\!10^{-6}$), batch size $8$, $5$ epochs, and $256K$ context length (full setup in Appendix~\ref{sec:appendix_setup}).

\paragraph{Results.}
Table~\ref{tab:main_results} shows that ``Qwen3-30B-A3B-SFT'' yields an AS of 0.371, a \textbf{2.67$\times$} gain over the base model ($0.139$) and approaching frontier-model performance ($0.410$). Sub.~CR and Samp.~CR surge to 48.8\% and 44.0\% from 11.8\% and 7.5\%, respectively. Similar gains in Qwen3-4B-SFT, alongside scaling-driven improvements in the 30B variant, collectively validate our training dataset.

This result---achieved with only ${\sim}$4K trajectories from a single teacher---suggests substantial headroom through scaling data collection, multiple teachers, and RL from $ES_t$ feedback.

\subsection{Case Studies}
\label{subsec:case-studies}

We analyze representative successes and failures to understand the capabilities and limitations of current LLMs. Unlike kernel-centric approaches, pass generation enables LLMs to discover \emph{graph-level rewrite rules}, i.e., pattern matchers paired with fused replacements that generalize across samples. Full implementations are provided in Appendix~\ref{sec:appendix_sparkle}.

\paragraph{Success Case 1: Roll+Slice Fusion (MaskFormer).}
TorchInductor decomposes \texttt{roll} into multiple \texttt{slice+cat} ops, launching 6 kernels for an 8-operator subgraph. The LLM recognizes the equivalence of \texttt{roll(shift=3) + slice[:128]} to index arithmetic ($\text{idx} = (S + i - \text{shift}) \bmod S$), replacing the entire subgraph with a single fused kernel (\textbf{3.02$\times$} speedup).

\paragraph{Success Case 2: Masked Mean Pooling (BGE-Reranker).}
TorchInductor fails to fuse a 7-op chain (\texttt{cast$\to$mul$\to$sum$\to$sum$\to$clamp$\to$div$\to$cat}), yielding a ${\sim}50\%$ slowdown vs.\ eager. The LLM identifies the masked mean pooling semantics and generates a single kernel that accumulates $\sum(\text{mask} \cdot \text{hidden})$ and $\sum(\text{mask})$ in FP32 registers (\textbf{2.90$\times$} speedup, bitwise-identical).

\begin{table}[!t]
\centering
\small
\setlength{\tabcolsep}{4pt}
\caption{\textbf{Sparkle Cases.} Speedups vs.\ Eager and Inductor on subgraphs where Inductor underperforms Eager.}
\label{tab:sparkle-cases}
\begin{tabular}{lccccc}
\toprule
\textbf{Model (Pattern)} & \textbf{Dtype} & \textbf{vs.\ Eager} & \textbf{vs.\ Inductor} & \textbf{Max Diff} & \textbf{Kernels} \\
\midrule
MaskFormer (Roll+Slice) & bf16 & \textbf{1.65$\times$} & \textbf{3.02$\times$} & $0.031$ & 6 $\to$ 1 \\
BGE-Reranker (Masked Pool) & bf16 & \textbf{1.50$\times$} & \textbf{2.90$\times$} & $0.0$ & 7 $\to$ 1 \\
\bottomrule
\end{tabular}
\end{table}

In both cases, the compiler loses high-level semantics after decomposing operations into primitives; the LLM's advantage is recognizing \emph{composite intent} and directly lowering to fused implementations.

\paragraph{Failure Modes.}
Analysis identifies three systematic bottlenecks in agent-driven optimization:
\textbf{(1)~Boundary Misalignment}---Agents often misjudge arithmetic intensity, causing inefficient fusion of low-compute operators (e.g., ReLU) or redundant re-implementation of vendor-optimized primitives (e.g., Conv2d) in Triton.
\textbf{(2)~Cost-Model Blindness}---Lacking hardware awareness (e.g., register pressure, SRAM capacity), agents employ static tiling across varying shapes, resulting in significant gaps from roofline performance.
\textbf{(3)~Semantic Disruption}---Local rewrites frequently break optimization chains by replacing standard patterns with opaque kernels, disabling critical features like FlashAttention-2 routing.
These issues indicate that PassBench poses open problems requiring hardware-aware reasoning beyond current LLM capabilities.

\section{Conclusion and Future Work}
\label{sec:conclusion}

We present \textbf{PassNet}, the first large-scale ecosystem for LLM-driven compiler pass generation, comprising: (1)~\textbf{PassNet-Dataset}, featuring $18K$ unique computational graphs derived from $100$K real-world models; and (2)~\textbf{PassBench}, a suite of $200$ curated long-tail tasks evaluated via the \textbf{Error-aware Speedup Score ($ES_t$)} with layered integrity defenses against LLM exploitation. Experiments demonstrate that PassBench is highly discriminative and unsaturated: while frontier models trail TorchInductor by 37\% in aggregate, individual passes achieve up to $3\times$ speedup---identifying \textit{consistency}, rather than capability, as the primary bottleneck. Notably, fine-tuning on ${\sim}4K$ PassNet trajectories yields a $2.67\times$ performance gain, approaching frontier-model levels and validating PassNet as essential infrastructure for advancing LLM-driven compiler optimization.

\paragraph{Limitations and Future Directions.}
Current experiments focus on fusible tasks, following a curriculum that progresses from simpler cases to more challenging classical subgraph tasks.
Accordingly, PassBench currently targets inference on a single GPU (NVIDIA A30), while generalization to training-loop optimizations, multi-device settings, and diverse hardware remains open.
The dataset is skewed toward CV and NLP workloads (90.6\% combined), which may limit coverage of emerging domains such as scientific simulation and generative models. Our anti-cheating defenses, while effective against observed exploits, cannot guarantee completeness against future adversarial strategies. 

Although evaluation graphs are sourced from public repositories, memorization risk is limited: pass generation requires producing executable pattern matchers and rewriters tailored to each graph structure, not reproducing code snippets; the multi-graph formulation further requires generalization across shapes and dtypes, reducing overfitting to specific instances.

Future directions include multi-device pass generation, extension to more complex tasks (e.g., classical subgraph tasks), integration of hardware cost models as auxiliary context, reinforcement learning from $ES_t$ feedback, and continual expansion of the dataset to underrepresented domains. The entire ecosystem is publicly available.

\newpage
\bibliographystyle{unsrtnat}
\bibliography{ref}

\newpage
\appendix

\section{Dataset Quality Constraints}
\label{sec:appendix_constraints}

A user’s model, wrapped by the \texttt{pass\_net.extract}, is symbolically traced to generate a standardized set of files. This set forms a complete PassNet graph, including the high-level IR of the computation graph (\texttt{model.py}), metadata for inputs and weights (\texttt{input\_meta.py}, \texttt{weight\_meta.py}), SHA-based graph hash for deduplication (\texttt{graph\_hash.txt}), and other components such as optional custom operator code.

We define five constraints applied to every computational graph in PassNet to ensure dataset quality and cross-platform compatibility:
\begin{itemize}[leftmargin=*,itemsep=2pt]
    \item \textbf{Runnable}: Each graph must execute forward propagation under the designated framework without syntax errors or crashes.
    \item \textbf{Serializable}: Each sample and metadata must be serializable into standard formats (e.g., JSON) and correctly de-serializable.
    \item \textbf{Decomposable}: The entire computational graph must be decomposable into multiple non-overlapping subgraphs, where each subgraph represents an independent optimization unit. This supports compiler backends in performing fusion, scheduling, and other optimization tasks.
    \item \textbf{Statically Analyzable}: Operator names, types, and dependencies must be statically extractable (e.g., via \texttt{torch.fx}) for structural traversal. This allows automated analysis tools to fully interpret operator semantics for structural traversal and pattern matching.
    \item \textbf{Custom Operator Accessible}: If a sample includes user-defined custom operators, the corresponding source code for these operators must be traceable and accessible in a modular form, ensuring reusability and integration across compiler environments.
\end{itemize}

\section{PassBench Sampling Details}
\label{sec:appendix_passbench_sampling}

\paragraph{Multi-dimensional Bucketing.}
Subgraphs are grouped into discrete buckets defined by three complementary dimensions:
\begin{itemize}[leftmargin=*,itemsep=2pt]
    \item \textbf{Operator Sequence}: The ordered operator-name list serves as an exact-match key.
    \item \textbf{Input Shape}: We apply logarithmic quantization $\lfloor \log_2(d)/4 \rfloor$ to each dimension $d$, where the scaling factor $4$ is empirically chosen to balance granularity and generalization.
    \item \textbf{Input Dtype}: Exact-match key for numerical precision.
\end{itemize}

\paragraph{Hierarchical Representative Grouping.}
Following bucketing, we employ a hierarchical strategy to select representative subgraphs and construct groups with varying cardinalities, each corresponding to a PassBench sample. The group size controls the optimization difficulty, as larger and more heterogeneous groups require transformations that generalize across a broader set of subgraphs.
\begin{itemize}[leftmargin=*,itemsep=2pt]
    \item \textbf{Intra-bucket Stratified Sampling}: Within each operator sequence, subgraphs are sampled at a fixed stride $\sigma$, and then organized into groups of size 1 and 3, capturing both single-operator cases and short compositional patterns.
    \item \textbf{Cross-shape Structural Aggregation}: For each unique operator sequence, one representative is aggregated with subgraphs sharing the same sequence across different input shape buckets.
    \item \textbf{Precision-aware Coverage}: Within each shape bucket, subgraphs with distinct data types (FP32, FP16, BF16) are aggregated to ensure numerical format coverage.
\end{itemize}

\section{Experimental Setup}
\label{sec:appendix_setup}

\subsection{Benchmark Evaluation}
Table~\ref{tab:eval_setup} summarizes the hardware and evaluation protocol used across all PassBench experiments.

\begin{table}[!htbp]
\centering
\small
\begin{tabular}{lc}
\toprule
\textbf{Category} & \textbf{Detail} \\
\midrule
GPU & NVIDIA A30, 24 GB, compute capability 8.0 \\
CUDA / cuDNN & 12.8 / 9.10.2 \\
PyTorch / Triton & 2.9.1+cu128 / 3.5.1 \\
Operating System & Ubuntu 24.04.1 LTS \\
\midrule
Evaluation Mode & Single-shot, temperature$=0$ \\
Warmup Runs & 20 \\
Timed Trials & 100 \\
Stability Criterion & Re-run if IQR $>$ 20\% of median \\
\bottomrule
\end{tabular}
\caption{\textbf{Evaluation environment and benchmark protocol.}}
\label{tab:eval_setup}
\end{table}

\subsection{Distillation and Post-training}
Table~\ref{tab:distill_setup} details the teacher-student configuration and SFT hyperparameters.

\begin{table}[!htbp]
\centering
\small
\begin{tabular}{lc}
\toprule
\textbf{Category} & \textbf{Detail} \\
\midrule
Teacher Agent & Claude-Sonnet-4.6 \\
Max Trajectory Steps & 50 per instance \\
Filtering Criterion & AS score $>$ 0.1 \\
Instances Processed & 4,476 (2 trials each) \\
Trajectories Retained & 3,899 \\
\midrule
Student Model & Qwen3-30B-A3B-Thinking-2507 \\
Initial Learning Rate & $2\times10^{-5}$ (cosine decay to $2\times10^{-6}$) \\
Batch Size & 8 \\
Epochs & 5 \\
Max Context Length & 262,144 \\
\bottomrule
\end{tabular}
\caption{\textbf{Distillation and post-training configuration.}}
\label{tab:distill_setup}
\end{table}

\section{Graph-level Interpretation of \texorpdfstring{$ES_t$}{ESt}}
\label{sec:appendix_es_geometric_mean}

In this section, we show that $ES_t$ can be equivalently expressed as a geometric mean of per-graph error-aware rectified speedups.

We first define $r_{t,i}$, the penalty factor for erroneous graphs that captures tolerance-dependent penalties. The per-graph error-aware rectified speedup $\hat{s}_{t,i}$ (defined in Section~\ref{sec:es-metric}) uses $r_{t,i}$ in its third case.

\begin{definition}[Penalty Factor]
\label{def:rectified-speedup}
For each erroneous graph $i$, let $c_i \in \{1,2,3\}$ denote its error category. Let $N_{\text{err}}$ be the number of erroneous graphs, and $N_c$ be the number of graphs with error category $c$. The \emph{penalty factor} is
\begin{equation}
r_{t,i} =
\begin{cases}
b, & t < c_i,\\[3pt]
1, & \text{otherwise},
\end{cases}
\end{equation}
where $b\in(0,1)$ is the base penalty.
\end{definition}

Let $\pi_c = N_c / N_{\text{err}}$ denote the fraction of error category $c$ among all erroneous graphs.

\begin{definition}[Error-aware Speedup Score $ES_t$]
The macro-level Error-aware Speedup Score $ES_t$ admits the following factored form:
\begin{equation}
ES_t = \alpha^{\lambda} \cdot \beta^{\lambda \eta (p+1)} \cdot \gamma_t^{1-\lambda}
\end{equation}
where $\gamma_t = b^{\sum_{c\in\{1,2,3\}} \pi_c \, \mathbf{1}(t<c)}$. Here $\alpha$ aggregates correct-and-fast subgraphs, $\beta$ penalizes correct-but-slow ones, and $\gamma_t$ accounts for errors under tolerance $t$.
\end{definition}

\begin{proposition}[Geometric mean of penalty factors]
\label{prop:equivalence-St-GMRS}
The aggregated penalty $\gamma_t$ equals the geometric mean of $\{r_{t,i}\}$ over all erroneous graphs:
\begin{equation}
\gamma_t = \left(\prod_{i=1}^{N_{\text{err}}} r_{t,i}\right)^{1/N_{\text{err}}}
\end{equation}
From the definition of $r_{t,i}$, each term contributes $b$ iff $t<c_i$:
\[
\prod_{i=1}^{N_{\text{err}}} r_{t,i}
 = \prod_{i=1}^{N_{\text{err}}} b^{\mathbf{1}(t<c_i)}
 = b^{\sum_{i=1}^{N_{\text{err}}}\mathbf{1}(t<c_i)}
\]
Grouping by error category $c$ gives
\[
\sum_{i=1}^{N_{\text{err}}}\mathbf{1}(t<c_i)
   = \sum_{c \in \{1,2,3\}}N_c\,\mathbf{1}(t<c)
\]
thus
\[
\prod_{i=1}^{N_{\text{err}}} r_{t,i}
   = b^{\sum_{c \in \{1,2,3\}}N_c\,\mathbf{1}(t<c)}
\]
Taking the $N_{\text{err}}$th root:
\[
\left(\prod_{i=1}^{N_{\text{err}}} r_{t,i}\right)^{1/N_{\text{err}}}
   = b^{\frac{1}{N_{\text{err}}}\sum_{c \in \{1,2,3\}}N_c\,\mathbf{1}(t<c)}
   = b^{\sum_{c \in \{1,2,3\}}\pi_c\,\mathbf{1}(t<c)}
\]
which is exactly the definition of $\gamma_t$.
\end{proposition}

\begin{proposition}[Geometric mean of per-graph speedups]
Since $\gamma_t$ is the geometric mean of $\{r_{t,i}\}$ for a given $t$, the macro-level metric $ES_t$ equals the geometric mean of per-graph error-aware rectified speedups:
\begin{equation}
ES_t = \left(\prod_{i=1}^{N}\hat{s}_{t,i}\right)^{1/N}
\end{equation}
This follows from the factored form: $\alpha$, $\beta$, and $\gamma_t$ are respectively the geometric means of the speedups from the correct-fast, correct-slow, and erroneous cases in $\hat{s}_{t,i}$. Their product therefore equals the geometric mean of all $\{\hat{s}_{t,i}\}$.
\end{proposition}

\section{Aggregated Speedup (AS) Weight Specification}
\label{sec:appendix_wt}

The Aggregated Speedup aggregates the tolerance-parameterized $ES_t$ into a single scalar via a normalized geometric mean (Section~\ref{sec:es-metric}). Given the weight function $W_t$, AS is defined as:
\begin{equation}
\mathrm{AS} = \prod_{t=-10}^{|E|+1} ES_t^{\,W_t / \sum_{s=-10}^{|E|+1} W_s}
\end{equation}

The weight function $W_t$ is defined as:
\begin{equation}
W_t =
\begin{cases}
  0.001 & \text{if } -10 \le t \le -6 \text{ or } t \ge |E|+1, \\
  1     & \text{if } -5 \le t \le -3, \\
  0.8^{t+3} & \text{if } -2 \le t \le |E|.
\end{cases}
\end{equation}

\paragraph{Design Rationale.}
The weight schedule reflects three regimes:
\begin{itemize}[leftmargin=*,itemsep=2pt]
    \item \textbf{Strict-correctness regime} ($t \in [-5, -3]$, $W_t = 1$): Production-level accuracy; full weight.
    \item \textbf{Relaxed regime} ($t \in [-2, |E|]$, $W_t = 0.8^{t+3}$): Exponential decay as correctness becomes easier to satisfy.
    \item \textbf{Extreme regime} ($t \le -6$ or $t \ge |E|+1$, $W_t = 0.001$): Near-zero weight to avoid distortion from ultra-strict or ultra-relaxed tolerances.
\end{itemize}

\section{Configuration of atol(t) and rtol(t)}
\label{sec:appendix_atol2rtol}

We perform log-linear interpolation between reference points (e.g., $\mathrm{atol}_{\mathrm{fp32}}(-5) = 10^{-5}$ and $\mathrm{atol}_{\mathrm{fp32}}(0) = 1$) such that $\mathrm{atol}(t), \mathrm{rtol}(t) = 10^{kt}$.

\begin{table}[!htbp]
\centering
\begin{minipage}[t]{0.49\textwidth}
\centering
\footnotesize
\begin{tabular}{cccc}
\toprule
\textbf{Data Type} & \textbf{atol(t)} & \textbf{atol}(-5) & \textbf{atol}(0)  \\ \midrule
bfloat16    &  $10^{t}$   & 1e-5   &  $1$    \\
float16/complex32    &  $10^{t}$   & 1e-5   &  $1$    \\
float32/complex64    &  $10^{t}$   & 1e-5   &  $1$    \\
float64/complex128    &  $10^{t\cdot7/5}$ & 1e-7   &  $1$    \\
others     &  $0.0$   & $0.0$         &  $0.0$  \\ \bottomrule
\end{tabular}
\caption{atol configuration (abbreviated)}
\end{minipage}
\hfill
\begin{minipage}[t]{0.49\textwidth}
\centering
\footnotesize
\begin{tabular}{cccc}
\toprule
\textbf{Data Type} & \textbf{rtol(t)} & \textbf{rtol}(-5) & \textbf{rtol}(0)  \\ \midrule
bfloat16     & $10^{t\cdot1.796/5}$         & 1.6e-2       & $1$    \\
float16/complex32     & $10^{t\cdot3/5}$         & 1e-3       & $1$    \\
float32/complex64     & $10^{t\cdot5.886/5}$         & 1.3e-6   & $1$  \\
float64/complex128     & $10^{t\cdot7/5}$         & 1e-7       & $1$    \\
others      & $0.0$       & $0.0$           & $0.0$  \\ \bottomrule
\end{tabular}
\caption{rtol configuration (abbreviated)}
\end{minipage}
\end{table}

\section{TorchInductor Default Pipeline Profiling}
\label{sec:appendix_inductor_profiling}

To empirically quantify the long-tail optimization gap discussed in Section~\ref{sec:intro}, we profile TorchInductor's default pass pipeline (\texttt{torch.compile(mode="default")}) on a large-scale subgraph corpus.

\paragraph{Setup.}
We extract $9{,}526$ subgraphs from $1{,}000$ community models via \texttt{torch.fx} graph tracing: 56.9\% from HuggingFace Transformers (BERT, T5, LLaVA, Qwen-VL, etc.) and 43.1\% from timm (ResNet, EfficientNet, ViT, FastViT, etc.). Subgraph node counts range from 4 to 63 (median 16). Experiments run on NVIDIA A30 GPUs (24\,GiB) with PyTorch 2.9.1 and CUDA 12.8. Each subgraph is benchmarked with $20$ warmup iterations and $100$ timed trials; we report median latencies and reject samples with unstable timing (IQR/median $> 20\%$).

\paragraph{Compilation Success.}
Of the $9{,}526$ subgraphs, 84.5\% compile successfully and pass correctness verification. The majority of failures (72.8\%) stem from GPU timing instability in the shared cloud environment (IQR/median exceeding our 20\% stability threshold); these samples are excluded as invalid measurements. The compiler-specific failure rate is 3.1\%, concentrated on non-standard architectures (LeViT, GCViT) with unconventional reshape and attention-head operations.

\paragraph{Kernel Speedup Distribution.}
Table~\ref{tab:inductor_kernel_dist} reports the kernel-level speedup distribution over the $8{,}021$ subgraphs with valid performance data.

\begin{table}[!htbp]
\centering
\small
\begin{tabular}{lr}
\toprule
Speedup Range & Fraction \\
\midrule
$< 1.0\times$ (degradation) & 8.3\% \\
$1.0$--$1.2\times$ (marginal) & 26.6\% \\
$1.2$--$2.0\times$ & 37.4\% \\
$\ge 2.0\times$ & 27.2\% \\
\bottomrule
\end{tabular}
\caption{\textbf{Kernel speedup distribution under TorchInductor's default pipeline.} Over one-third of subgraphs receive less than $1.2\times$ kernel acceleration.}
\label{tab:inductor_kernel_dist}
\end{table}

\paragraph{End-to-End Slowdown Analysis.}
Of the 43\% of subgraphs that exhibit E2E slowdowns, the majority are \emph{not} caused by kernel-level regression. Table~\ref{tab:e2e_badcase} decomposes the $3{,}479$ E2E bad cases by root cause.

\begin{table}[!htbp]
\centering
\small
\begin{tabular}{lrr}
\toprule
Root Cause & Fraction & Median E2E Speedup \\
\midrule
Small-graph fixed overhead ($<\!15$ nodes) & 61.5\% & $0.68\times$ \\
Fast-model dispatch overhead ($\ge\!15$ nodes, eager $<\!2$\,ms) & 18.7\% & $0.89\times$ \\
True kernel degradation & 18.3\% & $0.86\times$ \\
Other (marginal) & 1.5\% & $0.99\times$ \\
\bottomrule
\end{tabular}
\caption{\textbf{Root-cause decomposition of E2E slowdowns.} In 80.2\% of cases, kernels are actually faster (median $1.21\times$), but a fixed dispatch overhead of ${\sim}0.14$\,ms from \texttt{torch.compile} negates the gain.}
\label{tab:e2e_badcase}
\end{table}

Subgraph size is the primary determinant: the E2E bad rate drops from 85.8\% for subgraphs with 1--5 nodes to 13.0\% for those with 30--50 nodes (median subgraph size: 9 nodes for bad cases vs.\ 29 for good cases). This confirms that the E2E overhead is largely a framework infrastructure cost orthogonal to pass-level optimization quality. True kernel degradation accounts for only 18.3\% of E2E bad cases and is concentrated on FastViT-family architectures with non-standard attention mechanisms.

\paragraph{Numerical Precision.}
Among the $8{,}027$ correctly compiled subgraphs, 39.7\% produce bitwise-identical outputs and 81.5\% exhibit maximum absolute difference below $10^{-6}$. A small fraction (4.6\%) shows deviations $\ge 10^{-3}$, typically correlated with aggressive fusion strategies that yield higher speedups (mean kernel speedup $3.03\times$ vs.\ $2.09\times$ for precise cases).

\paragraph{ES($t$) Scores.}
Evaluating TorchInductor under our $ES_t$ metric yields a G-Mean Speedup of $0.846$ and an AS Score of $0.706$ (Table~\ref{tab:main_results}), confirming substantial room for improvement when correctness and stability are jointly considered.

\section{Detailed Sparkle Case Analysis}
\label{sec:appendix_sparkle}

This appendix provides the complete pass files for the sparkle cases discussed in Section~\ref{sec:experiments}.

\subsection{Case 1: Index-Arithmetic Roll+Slice Fusion (MaskFormer)}

The MaskFormer pass targets an 8-operator subgraph in MaskFormer's pixel decoder. Its operator sequence is \texttt{contiguous$\to$view$\to$roll$\allowbreak$$\to$slice$\to$contiguous$\allowbreak$$\to$view$\to$add$\to$layer\_norm}; under default compilation, this launches 6 separate kernels. The LLM recognizes that \texttt{roll(shift=3)+slice[:128]} on a $[S,S,C]$ tensor can be replaced by direct index arithmetic $\text{input\_i}=(S+i-\text{shift})\bmod S$, and additionally fuses the layer-norm reduction via shared-memory accumulation. Crucially, the replacement produces \textbf{two outputs} (add result and layer-norm result) in a single kernel.

The pass is split into a \textbf{shared lowering module} and a \textbf{shape-specific pattern pass}. The shared module contains the CUDA kernel, the inline loader registration, and a routing dispatch that parameterizes the kernel for three Swin-Transformer stages (D=96, 192, 384). All shape-specific passes import the same shared module and differ only in their \texttt{pattern} and \texttt{replacement\_args}, demonstrating the reusability of a single lowering across multiple FX pattern instances.

\textbf{Shared lowering module (\texttt{shared\_fused\_roll\_add\_ln.py} with inline CUDA):}
\begin{verbatim}
import torch
from torch.utils.cpp_extension import load_inline

_cuda_src = r"""
#include <torch/extension.h>
#include <cuda_runtime.h>

template<typename scalar_t>
__global__ void fused_roll_slice_add_layernorm_kernel(
    const scalar_t* __restrict__ input_6d,
    const scalar_t* __restrict__ residual,
    const scalar_t* __restrict__ weight,
    const scalar_t* __restrict__ bias,
    scalar_t* __restrict__ out_add,
    scalar_t* __restrict__ out_ln,
    const int R, const int C, const int S, const int SC, const int shift
) {
    const int row = blockIdx.x;
    if (row >= R) return;

    extern __shared__ float smem[];
    const int tid = threadIdx.x;

    // Index arithmetic: roll(shift,dims=[0,1]) on [S,S,C] then slice [:SC,:SC,:]
    const int i = row / SC;
    const int j = row % SC;
    const int input_i = (S + i - shift) % S;
    const int input_j = (S + j - shift) % S;

    // Pass 1: accumulate mean of (residual + rolled_input) via shared memory
    float local_sum = 0.0f;
    for (int col = tid; col < C; col += blockDim.x) {
        float rolled = static_cast<float>(
            input_6d[input_i * S * C + input_j * C + col]);
        float res = static_cast<float>(residual[row * C + col]);
        local_sum += res + rolled;
    }
    smem[tid] = local_sum;
    // shared-memory reduction to mean ...
    const float mean = smem[0] / static_cast<float>(C);

    // Pass 2: accumulate variance
    float local_var = 0.0f;
    for (int col = tid; col < C; col += blockDim.x) {
        float rolled = static_cast<float>(
            input_6d[input_i * S * C + input_j * C + col]);
        float res = static_cast<float>(residual[row * C + col]);
        float diff = (res + rolled) - mean;
        local_var += diff * diff;
    }
    smem[tid] = local_var;
    // shared-memory reduction to variance ...
    const float inv_std = rsqrtf(smem[0] / static_cast<float>(C) + 1e-5f);

    // Pass 3: write both outputs
    for (int col = tid; col < C; col += blockDim.x) {
        float rolled = static_cast<float>(
            input_6d[input_i * S * C + input_j * C + col]);
        float res = static_cast<float>(residual[row * C + col]);
        float added = res + rolled;
        out_add[row * C + col] = static_cast<scalar_t>(added);
        float normed = (added - mean) * inv_std;
        out_ln[row * C + col] = static_cast<scalar_t>(
            normed * static_cast<float>(weight[col])
            + static_cast<float>(bias[col]));
    }
}
// Host wrapper omitted for brevity; allocates out_add/out_ln,
// chooses threads via next-power-of-two heuristics, and launches.
"""

_ext = load_inline(
    name="fused_roll_slice_add_layernorm_ext",
    cuda_sources=_cuda_src,
    functions=["fused_roll_slice_add_layernorm_cuda"],
    verbose=False,
)


@torch.fx.wrap
def fused_roll_slice_add_layernorm_dispatch(in_3, in_2, in_1, in_0, route):
    if route == "D96":
        return _ext.fused_roll_slice_add_layernorm_cuda(
            in_3, in_2, in_1, in_0, 133, 128, 3, 1e-5)
    elif route == "D192":
        return _ext.fused_roll_slice_add_layernorm_cuda(
            in_3, in_2, in_1, in_0, 70, 64, 3, 1e-5)
    elif route == "D384":
        return _ext.fused_roll_slice_add_layernorm_cuda(
            in_3, in_2, in_1, in_0, 35, 32, 3, 1e-5)
    else:
        raise ValueError(f"Unknown route: {route}")
\end{verbatim}

\textbf{Shape-specific pattern pass (\texttt{FusedRollSliceAddLayerNorm\_96.py}, D=96 stage):}
\begin{verbatim}
import torch
from pass_dir.shared_fused_roll_add_ln import fused_roll_slice_add_layernorm_dispatch


def pattern(in_0, in_1, in_2, in_3):
    tmp_2 = in_3.contiguous()
    tmp_3 = tmp_2.view(-1, 133, 133, 96)
    tmp_4 = torch.roll(tmp_3, shifts=(3, 3), dims=(1, 2))
    tmp_5 = tmp_4[(slice(None, None, None), slice(None, 128, None),
                   slice(None, 128, None), slice(None, None, None))]
    tmp_6 = tmp_5.contiguous()
    tmp_7 = tmp_6.view(1, 16384, 96)
    tmp_8 = in_2 + tmp_7
    tmp_9 = torch.nn.functional.layer_norm(tmp_8, (96,), in_1, in_0, 1e-05)
    return (tmp_8, tmp_9)


def replacement_args(in_0, in_1, in_2, in_3):
    return (in_3, in_2, in_1, in_0, "D96")


def replacement_func():
    return fused_roll_slice_add_layernorm_dispatch
\end{verbatim}

\subsection{Case 2: Pattern-Driven Masked Mean Pooling (BGE-Reranker)}

The BGE-Reranker pass targets a 7-operator subgraph in the sentence-embedding module. Its operator sequence is \texttt{cast$\to$mul$\allowbreak$$\to$sum$\to$sum$\allowbreak$$\to$clamp$\to$div$\allowbreak$$\to$cat}; under default compilation, this launches 7 separate kernels because the compiler treats the two reductions independently. The LLM recognizes the semantic intent as ``masked mean pooling'' and generates a rewrite that fuses all operators into a single kernel, accumulating $\sum(\text{mask}\cdot\text{hidden})$ and $\sum(\text{mask})$ simultaneously in float32 registers.

\textbf{Shared lowering module (\texttt{shared\_cuda\_bge.py} with inline CUDA):}
\begin{verbatim}
import torch
from torch.utils.cpp_extension import load_inline

_cuda_src = r"""
#include <torch/extension.h>
#include <cuda_runtime.h>

template<typename hidden_t>
__global__ void masked_mean_pooling_kernel(
    const float* __restrict__ mask,
    const hidden_t* __restrict__ hidden,
    float* __restrict__ output,
    const int S, const int D
) {
    const int bid = blockIdx.x;
    const int d_idx = blockIdx.y * blockDim.x + threadIdx.x;
    if (d_idx >= D) return;

    const float* mask_batch = mask + bid * S * D;
    const hidden_t* hidden_batch = hidden + bid * S * D;

    float acc_val = 0.0f;
    float acc_mask = 0.0f;
    for (int s = 0; s < S; ++s) {
        int offset = s * D + d_idx;
        acc_val += static_cast<float>(hidden_batch[offset]) * mask_batch[offset];
        acc_mask += mask_batch[offset];
    }
    output[bid * D + d_idx] = acc_val / (acc_mask > 1e-9f ? acc_mask : 1e-9f);
}
// Host wrapper omitted for brevity; converts mask to float32,
// allocates [B,D] float32 output, and launches with dim3(B, ceil(D/128)).
"""

_ext = load_inline(
    name="fused_masked_mean_pooling_cuda_ext",
    cuda_sources=_cuda_src,
    functions=["fused_masked_mean_pooling_cuda"],
    verbose=False,
)


@torch.fx.wrap
def fused_masked_mean_pooling(in_0, in_1):
    return _ext.fused_masked_mean_pooling_cuda(in_0, in_1)
\end{verbatim}

\textbf{Pattern pass (\texttt{FuseMaskedMeanPooling.py}):}
\begin{verbatim}
import torch
from pass_dir.shared_cuda_bge import fused_masked_mean_pooling


def pattern(in_0, in_1):
    tmp_0 = in_0.to(torch.float32)
    tmp_1 = in_1 * tmp_0
    tmp_2 = torch.sum(tmp_1, 1)
    tmp_3 = tmp_0.sum(1)
    tmp_4 = torch.clamp(tmp_3, min=1e-09)
    tmp_5 = tmp_2 / tmp_4
    tmp_6 = torch.cat([tmp_5], 1)
    return tmp_6


def replacement_args(in_0, in_1):
    return (in_0, in_1)


def replacement_func():
    return fused_masked_mean_pooling
\end{verbatim}

\textbf{Key implementation notes.} The kernel performs two sequential reductions (\texttt{acc\_val} and \texttt{acc\_mask}) in float32 over the sequence dimension $S$, matching the eager-mode cast-to-float32 semantics. The output shape is $[B, D]$, where $B$ is the batch size and $D$ is the hidden dimension. Because the eager pattern includes a no-op \texttt{cat} (concatenating a single tensor along dim 1), the replacement preserves the original output shape without additional reshaping. The same rewrite rule applies to varying sequence lengths and hidden dimensions without recompilation.

\end{document}